\documentclass{ieeetj}
\usepackage{cite}
\usepackage{amsmath,amssymb,amsfonts}
\usepackage{graphicx,color}
\usepackage{textcomp}
\usepackage{xcolor}
\usepackage{hyperref}
\usepackage{xspace}
\usepackage{booktabs}
\usepackage{fontawesome5}
\hypersetup{hidelinks=true}
\usepackage{algorithm,algorithmic}
\def\BibTeX{{\rm B\kern-.05em{\sc i\kern-.025em b}\kern-.08em
    T\kern-.1667em\lower.7ex\hbox{E}\kern-.125emX}}
\AtBeginDocument{\definecolor{tmlcncolor}{cmyk}{0.93,0.59,0.15,0.02}\definecolor{NavyBlue}{RGB}{0,86,125}}

\def\OJlogo{\vspace{-4pt}Preprint.}

\def\authorrefmark#1{\ensuremath{^{\textbf{#1}}}}

\usepackage[capitalize,nameinlink,noabbrev]{cleveref}

\newcommand{\eg}{e.g.,\xspace}
\newcommand{\ie}{i.e.,\xspace}
\newcommand{\etal}{et al.\xspace}

\newcommand{\method}{EgoM\textsuperscript{3}\xspace}

\newcommand{\pampjpe}{PA-MPJPE\xspace}
\newcommand{\wampjpe}{WA-MPJPE\textsubscript{100}\xspace}

\newcommand{\rampjpe}{RA-MPJPE\textsubscript{100}\xspace}

\begin{document}
\bstctlcite{IEEEexample:BSTcontrol}

\markboth{\method: Multi-View Multi-Human Mesh Reconstruction from Egocentric Videos}{Hokari {et al.}}

\title{\method: Multi-View Multi-Human Mesh Reconstruction from Egocentric Videos}

\author{Yamato Hokari\authorrefmark{1}, Zhifan Zhu\authorrefmark{2}, Ryosuke Hori\authorrefmark{1}, Hiroyuki Deguchi\authorrefmark{1}, and Hideo Saito\authorrefmark{1}}
\affil{Keio University, Yokohama, Japan}
\affil{University of Bristol, Bristol, UK}
\corresp{Corresponding author: Hideo Saito (email: hs@keio.jp).}
\authornote{\large\href{https://shigurex.github.io/egom3}{\textcolor{NavyBlue}{\faGlobe\enspace\textbf{Project Page: https://shigurex.github.io/egom3}}}}

\begin{abstract}
We address the problem of reconstructing a group of people using only egocentric videos,
without any stationary or third-person cameras.
Several participants wear egocentric cameras and perform a group activity while moving freely.
The participants cannot see their own bodies, but can see the others,
and we aim to reconstruct the body meshes of all participants in a shared coordinate system. 
This setting is challenging because the cameras are constantly moving, the same person must be matched across views, and participants may temporarily go out of view.
We address these challenges by proposing a three-stage method, \method.
Given RGB videos, 
\method first estimates the camera poses and per-view human meshes in a shared coordinate system.
Second, \method matches body meshes across views and time, and associates each participant to their camera, both by considering geometric consistency.
Third, with the associated meshes as guidance,
\method uses a pretrained diffusion model to produce a single, complete motion for each participant.
We evaluate \method on 34 sequences from the EgoHumans dataset,
using the individual-level metric (\wampjpe) and a new group-level metric.
Our method reduces \wampjpe by 34.1\% and the group-level error by 23.3\% compared to the baseline that conditions EgoAllo on the camera trajectories estimated by Pi3X.
\end{abstract}

\begin{IEEEkeywords}
Egocentric Cameras, Human Pose Estimation, Motion Completion, Multiple Views
\end{IEEEkeywords}

\maketitle

\section{INTRODUCTION}
\label{sec:intro}

\begin{figure*}[t]
  \centering
  \includegraphics[width=\linewidth]{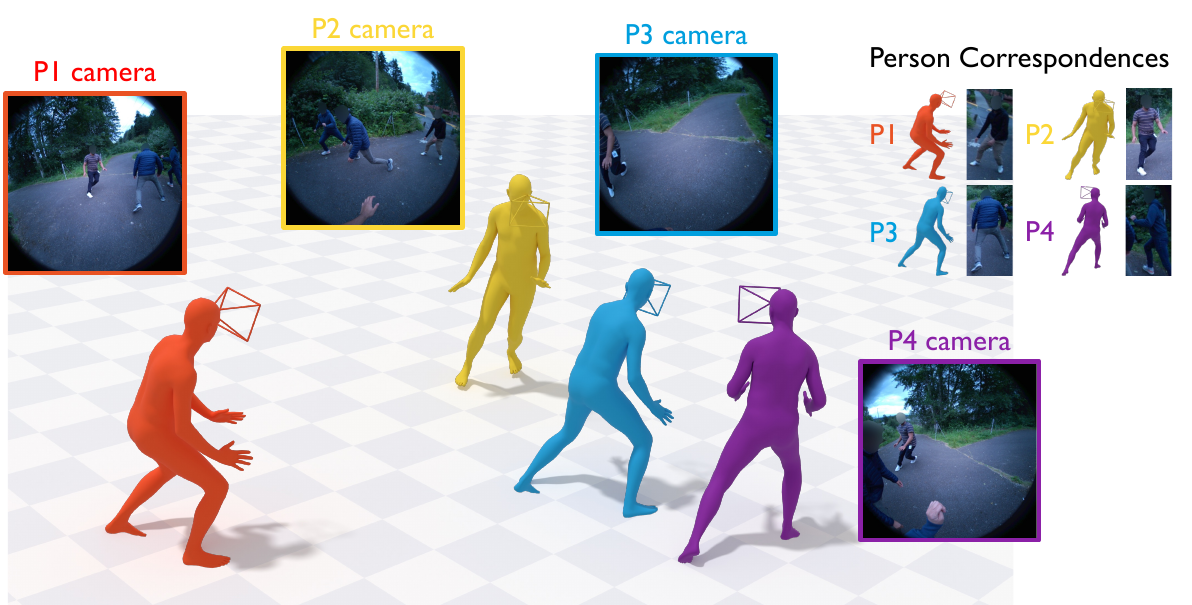}
  \caption{
    The EgoMocap problem.
    Four participants wear egocentric cameras that observe
    one another but not their own bodies.
    For example, P1 observes P2, P3, and P4 from left to right in the egocentric view,
    while P1 is visible only to P2.
    EgoMocap aims to reconstruct each participant's mesh
    in a shared coordinate system, 
    despite heavy occlusions and incomplete observations.
    The meshes are reconstructed by our method.
    This figure shows one timestep from the sequence.
    \emph{Top-right}: Corresponding image observations
    from the egocentric views.
  }
  \label{fig:teaser}
\end{figure*}

In team sports, collaborative work, and everyday social
activities, people move, turn, and interact with one another as a group.
Reconstructing the 3D body meshes of all members in these group activities 
facilitates understanding of collaboration and social interaction, 
and has potential applications in sports and embodied AI.
To reconstruct their meshes, 
existing approaches generally assume the people are captured by third-person cameras,
either with stationary multi-camera rigs~\cite{panoptic2015,khirodkar2023egohumans} or moving third-person views~\cite{ye2023slahmr,newell2025comotion,chen2026human3r,kim2026harmoni}.
However, when participants all wear egocentric cameras, the cameras themselves offer a unique opportunity to reconstruct human meshes: participants naturally appear in each other's view as they move and engage in the activity.

In this work, 
we study the problem of reconstructing a group of participants using only their egocentric videos.
Given synchronized egocentric videos from multiple participants,
the goal is to reconstruct an SMPL-X~\cite{smplx2019} mesh for each one of them
in a shared coordinate system.
We refer to this problem as \emph{EgoMocap} and
illustrate this in~\cref{fig:teaser}.

While prior egocentric motion capture methods exist, 
they primarily reconstruct a single wearer and rely on body-facing cameras~\cite{xu2019mo2cap2} 
or additional hand tracking sensors~\cite{jiang2022avatarposer,jiang2024egoposer}.
Nakamura \etal~\cite{nakamura2017} pioneered exploiting mutual observation in the two-person case, 
and concurrently, Shin \etal~\cite{shin2026everybody} address a similar multi-person setting as EgoMocap.
However, Shin \etal require camera trajectories that are already given in a shared coordinate system, whereas we assume only RGB videos as input.
Solving EgoMocap from RGB videos alone presents three challenges:
(1) the cameras move with their wearers, and neither
the camera poses nor a common coordinate frame is given;
(2) the same person can be observed in different cameras, but observations of this person must be associated across views and time and linked to the camera they wear;
and 
(3) participants can be temporarily occluded or out of view,
causing direct reconstruction to miss their motion across frames.

To address these challenges, we propose \method (\textbf{M}ulti-view \textbf{M}ulti-human \textbf{M}esh), a three-stage method
that reconstructs the meshes of all participants from egocentric RGB videos alone.
First, inspired by HARMONI~\cite{kim2026harmoni}, 
\method employs a frozen scene encoder~\cite{pi3_2026} to extract scene features and estimate camera poses in a shared coordinate system, 
and a frozen human encoder~\cite{baradel2024multihmr} to extract features for each detected person.
A lightweight prediction head then fuses these features to estimate
an SMPL-X mesh for each detected person in each view.

Second, we transform the estimated meshes into the shared
coordinate system using the estimated camera poses.
We compute matching costs focusing on the geometric consistency
of body joints and  head poses.
These costs are used to match detections across views
at the same timestep and across time,
grouping meshes of the same person together.
We then associate each group with its wearer's camera
by measuring the geometric consistency between the camera pose and the
mesh's head pose.

Finally, 
we condition a pretrained motion diffusion model,
EgoAllo~\cite{yi2025egoallo}, on the estimated camera trajectories
and guide DDIM sampling~\cite{song2020denoising}
with the body poses from the above stage.
These camera trajectories are available even when the body is unobserved,
and the diffusion guidance enables complete and temporally coherent motion.

We evaluate \method on 34 sequences from the EgoHumans dataset~\cite{khirodkar2023egohumans}.
We further introduce a metric to evaluate global reconstruction quality at the group level.
Compared with baselines that directly combine off-the-shelf models,
our method achieves superior global reconstruction accuracy and motion quality.

Our contributions are threefold:
(1) We formalize \emph{EgoMocap}, a multi-person mesh
reconstruction problem using only participants' egocentric RGB videos.
(2) We propose a three-stage method \method to solve the problem, via
per-view mesh reconstruction,
cross-view ReID,
and diffusion-guided motion completion.
(3) We introduce \rampjpe, a group-level metric that evaluates spatiotemporal reconstruction accuracy across all participants jointly, and demonstrate consistent gains in global accuracy and motion quality on EgoHumans.
\section{RELATED WORK}
\label{sec:related}

\noindent\textbf{Monocular Human Mesh Recovery (HMR).}
Based on the input viewpoints,
monocular HMR methods can be divided into two categories: 
third-person based and egocentric based.
Most prior monocular HMR methods focus on third-person based input:
the early third-person method~\cite{kanazawa2018hmr} directly regresses
SMPL~\cite{loper2015smpl} parameters from a single image, and later methods advance in
transformer backbones and large-scale real and synthetic
data~\cite{goel2023humans,black2023bedlam,khirodkar2024harmony4d}.
Recent methods incorporate
perspective-aware camera estimation~\cite{patel2025camerahmr},
multi-person whole-body reconstruction~\cite{baradel2024multihmr,muller2025hsfm,kim2026harmoni},
and videos~\cite{ye2023slahmr, shin2024wham,shen2024gvhmr, wang2024tram, newell2025comotion, li2025genmo,chen2026human3r,on2026multi}.
In practice, these third-person methods assume stationary cameras or moving cameras only for observing,
while we focus on using egocentric cameras to reconstruct the wearers themselves.

Regarding egocentric based input,
most prior methods focus on downward-facing fisheye cameras~\cite{xu2019mo2cap2} or 
additional hand tracking signals~\cite{jiang2022avatarposer}.
A few methods also explored using motion priors to generate body poses~\cite{jiang2022avatarposer, li2023ego, guzov2025hmd2, yi2025egoallo, Hatano2025EgoH4}.
However, these methods focus on a single human and do not reconstruct multiple humans jointly.

Within ego-based methods, we note the following works closest to ours.
Nakamura \etal~\cite{nakamura2017} is a pioneering work that explores a mutual motion capture focusing on the two-person case. This work requires additional head-mounted displays.
The other two methods, EgoExoMoCap~\cite{jiang2026egoexomocap} and Shin \etal~\cite{shin2026everybody}, are concurrent work that both take as input multiple participants' videos as in our setting.
However, EgoExoMoCap only produces one person's mesh;
Shin \etal indeed produces all human meshes, 
but assumes participants' camera poses already given in a shared coordinate system.
In contrast, our method directly operates on RGB videos, 
without requiring external localizations.
Furthermore, prior evaluation relies solely on the individual metrics while we contribute a group-level metric for the multiple person setup. 

\noindent\textbf{Diffusion guidance for human motion.}
Diffusion models provide effective priors for human
motion~\cite{tevet2023human}, and inference-time
guidance incorporates external constraints without
retraining~\cite{bansal2023universal,chung2023diffusion}.
GMD~\cite{karunratanakul2023guided} enables spatially
constrained motion generation,
RoHM~\cite{zhang2024rohm} reconstructs motion from
incomplete observations through guided sampling, and
CondMDI~\cite{cohan2024flexible} learns keyframe-conditioned
motion in-betweening.
EgoAllo~\cite{yi2025egoallo} conditions motion on head
trajectories and incorporates hand observations through
guidance.
Using EgoAllo, our method  incorporates body pose
observations from other participants' cameras,
temporal smoothness, and foot-contact constraints
into sampling to recover coherent motion under
incomplete observations.
\section{METHOD}
\label{sec:method}

\begin{figure*}[t]
  \centering
  \includegraphics[width=\linewidth]{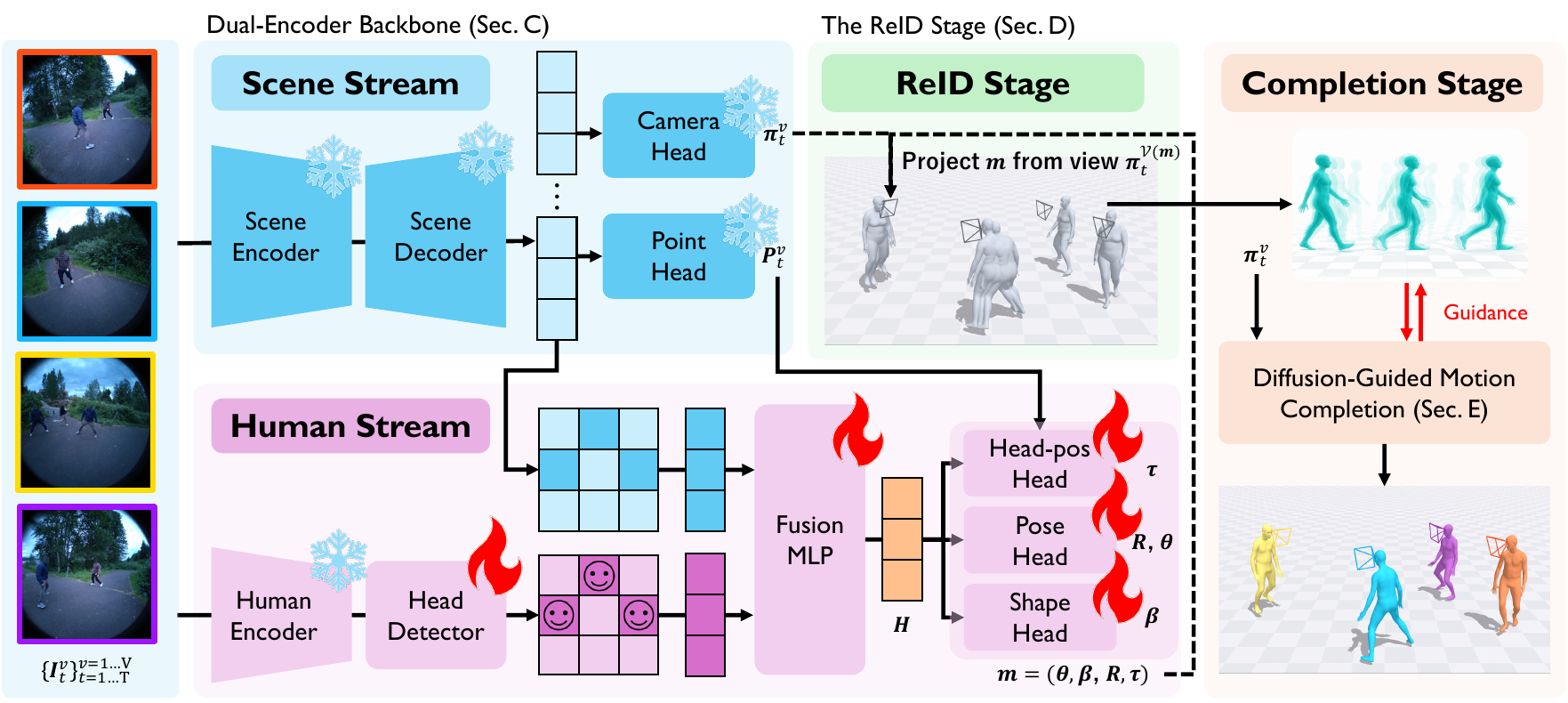}
  \caption{Overview of \method.
    The dual-encoder backbone (\cref{sec:method:backbone}) processes each view using two frozen encoders.
    The scene stream estimates camera poses $\boldsymbol{\pi}_t^v$
    and point maps $\boldsymbol{P}_t^v$,
    while the human stream detects visible people.
    Features from both streams are fused into person tokens
    $\boldsymbol{H}$, which trainable heads decode into
    head positions $\boldsymbol{\tau}$,
    root rotations $\boldsymbol{R}$,
    body poses $\boldsymbol{\theta}$,
    and shapes $\boldsymbol{\beta}$.
    The ReID stage (\cref{sec:method:assoc}) associates
    meshes across views and time, grouping them by identity
    and matching each group to its wearer's camera.
    Finally, the completion stage (\cref{sec:method:diffusion})
    uses a pretrained motion diffusion model conditioned on
    camera trajectories and guided by the associated meshes
    to reconstruct temporally complete motion.
    Smiling faces mark the detected heads.
    Snowflakes indicate frozen modules, and flames indicate
    trained modules.}
  \label{fig:overview}
\end{figure*}

\subsection{EGOMOCAP PROBLEM FORMULATION}
\label{sec:method:settings}

We assume $V$ participants wearing $V$ egocentric cameras,
and we aim to reconstruct the body mesh of each participant.
The full body of each person is not visible from their own view, 
but may be observed in one or more other participants' camera views.
Since each camera is worn by one participant, each mesh must also be associated with its wearer's camera.

\noindent\textbf{Input and Output.} 
Given synchronized videos $\{\boldsymbol{I}_t^v\}_{t=1...T}^{v=1...V}$ from $V$ egocentric cameras over $T$ timesteps, the goal of EgoMocap is to reconstruct all participants' meshes in a shared coordinate system.
At each timestep, EgoMocap expects precisely $V$ meshes in the output.
For each participant $v$ at timestep $t$,
its body mesh is represented using the SMPL-X~\cite{smplx2019} parameterization:
body articulation $\boldsymbol{\theta}_t^v\in\mathbb{R}^{52\times3}$, 
body shape $\boldsymbol{\beta}_t^v\in\mathbb{R}^{10}$, 
root rotation $\boldsymbol{R}_t^v\in\mathrm{SO}(3)$, 
and head position $\boldsymbol{\tau}_t^v\in\mathbb{R}^3$. 
Each mesh is also associated with a camera pose $\boldsymbol{\pi}_t^v \in \mathrm{SE}(3)$ that corresponds to the egocentric camera worn by that participant.
$\boldsymbol{R}_t^v$,  $\boldsymbol{\tau}_t^v$ and  $\boldsymbol{\pi}_t^v$ are expressed in the shared world coordinate system.

\subsection{OVERVIEW OF EGOM\textsuperscript{3}}

We first present an overview of \method and define notations.
Since each image $\boldsymbol{I}_t^v$ contains at most $V-1$ participants
and there are $V$ images per timestep,
a length-$T$ input contains at most $(V-1) \times V \times T $ in-view participants.
In practice, a sequence is processed in temporal chunks of $T_c$ consecutive timesteps,
whose outputs are chained across the full sequence of length $T$.

\method first estimates all observed humans' meshes,  using a dual-encoder backbone.
At each timestep $t$, the backbone produces $V$ sets of meshes, one for each view.
We denote the union of these $V$ sets as $M_t$,
\eg with $V=4$,

\begin{equation}
\resizebox{0.88\linewidth}{!}{$
M_t =\{
\underbrace{\vphantom{m_{|M_t|}} m_1, m_2, \dots}_{v=1} \,
\underbrace{\vphantom{m_{|M_t|}} \dots, m_i, \dots}_{v=2} \,
\underbrace{\vphantom{m_{|M_t|}} \dots}_{v=3} \,
\underbrace{\vphantom{m_{|M_t|}} \dots, m_{|M_t|}}_{v=4} \, \},
$}
\label{eq:Mt}
\end{equation}
where $m_1, m_2, \ldots, m_i, \ldots, m_{|M_t|}$ are estimated meshes.

We further use $\mathcal{V}(\cdot)$ to refer to the source view of each mesh, \eg$\mathcal{V}(m_i) = 2$ in the above equation.
It is possible that the set of a view $v$ may not contain any meshes, when no person is present in that view.
Note that the number of meshes $|M_t|$ varies across timesteps.
The sets $M_1, M_2, \ldots M_t,\ldots M_{T_c}$ are dual-encoder outputs.
In addition, the dual-encoder outputs camera poses $\{\boldsymbol{\pi}^v_t\}^{v=1 \ldots V}_{t=1 \ldots T_c}$.

The next ReID stage takes the above mesh sets 
as input and performs association. Association is performed across views within each
$M_t$, and across timesteps within each view.
The ReID stage outputs $V$ groups of meshes, where each group consists of the same person's meshes from different views.
In other words, $M_t$ is re-partitioned according to the mesh--person correspondences.

Lastly, the completion stage takes as input the camera poses
$\boldsymbol{\pi}^v_t$ from the dual-encoder and the mesh groups from the ReID stage, and combines them using a pretrained
motion diffusion model. The camera poses are the head trajectory of each
wearer, on which we condition the diffusion model, and the meshes guide the diffusion sampling.
This produces a temporally complete and smooth mesh for every participant,
including the timesteps when no other camera observes the participant.
\method is overviewed in \cref{fig:overview}. 

\subsection{DUAL-ENCODER BACKBONE}
\label{sec:method:backbone}

Our backbone design follows HARMONI~\cite{kim2026harmoni}.
The input $\{\boldsymbol{I}_t^v\}_{t=1\ldots T_c}^{v=1\ldots V}$ is fed into two separate feature extractors:
Pi3X~\cite{pi3_2026} for the scene stream and Multi-HMR~\cite{baradel2024multihmr} for the human stream.
The scene stream produces the camera poses and point maps, whereas the human stream provides the per-person features from which the meshes are decoded.

The scene stream Pi3X~\cite{pi3_2026} employs an encoder-decoder architecture to embed each image $\boldsymbol{I}_t^v$ into a list of tokens.
Pi3X subsequently processes these tokens with a camera head and a point head, producing a per-image camera pose $\boldsymbol{\pi}^v_t$ and a per-image point map $\boldsymbol{P}^v_t$.
The output camera poses $\{\boldsymbol{\pi}^v_t\}^{v=1\ldots V}_{t=1\ldots T_c}$ share the same (world-frame) coordinate system,
as a result of the joint processing of all $V \times T_c$ images within Pi3X.

For the human stream,
the Multi-HMR~\cite{baradel2024multihmr} human encoder extracts features independently for each image.
This produces a 2D grid of features for each image, with one feature vector per slot in the grid.
Given this grid, a lightweight MLP head detector outputs a scalar confidence for each slot.
After a $3\times3$ non-maximum suppression over the grid,
we keep the slots whose confidences are at least $\mathrm{thr} = 0.3$; these identify the detected heads.
A second lightweight MLP regresses a sub-slot offset that refines the image location $\boldsymbol{u}_i$ of each detected head.
\cref{fig:overview} shows the 2D grid with detected heads marked as smiling faces.

For each detected head, we estimate the body pose of the corresponding person.
Following HARMONI~\cite{kim2026harmoni}, we combine the human feature with the scene feature at the
same detected location.
To this end, we reshape the list of scene tokens into a 2D grid with one token per slot,
and sample this grid at the refined head location $\boldsymbol{u}_i$ by bilinear
interpolation to obtain the desired scene feature.

Given the scene feature and the human feature, 
a fusion MLP concatenates them as input and produces a person token $\boldsymbol{H}_{i}$.
We then use separate network heads to obtain the body articulation
$\boldsymbol{\theta}_{i}$, body shape
$\boldsymbol{\beta}_{i}$, camera-frame root rotation $\boldsymbol{R}_{i}$, and
camera-frame head position $\boldsymbol{\tau}_{i}$.
The head-position head reads the scene stream's
point map at the detected head and predicts a depth correction~\cite{kim2026harmoni}.
The estimates
$(\boldsymbol{\theta}_i,\boldsymbol{\beta}_i,\boldsymbol{R}_i,\boldsymbol{\tau}_i)$
together form the mesh $m_i$ in \cref{eq:Mt}, and are lifted to the shared coordinate system
with the camera pose $\boldsymbol{\pi}_t^{\mathcal{V}(m_i)}$ of the view that observed it.

\subsection{THE REID STAGE}
\label{sec:method:assoc}

Since the mesh estimates do not carry identity information,
we resolve this ambiguity in four steps:
(1) matching meshes across views at the same timestep,
(2) matching meshes across consecutive timesteps,
(3) grouping the accepted matches across views and time,
and (4) associating each group with an egocentric camera.

\noindent\textbf{Matching across views.}
For each view $v$, we define the subset of meshes
originating from view $v$ as
\begin{equation}
M_t^v
=
\left\{
m_i \in M_t
\mid
\mathcal{V}(m_i)=v
\right\}.
\label{eq:per-view-mesh-set}
\end{equation}

Since two meshes from the same view at the same timestep
must represent different people, we consider only meshes
from distinct views $a$ and $b$.
For each pair $m_i\in M_t^a$ and $m_j\in M_t^b$,
we define a person-to-person matching cost as
\begin{equation}
\mathcal{C}_{\mathrm{p2p}}:
M_t^a \times M_t^b
\rightarrow \mathbb{R},
\qquad
(m_i,m_j)
\mapsto
\mathcal{C}_{\mathrm{p2p}}(m_i,m_j).
\label{eq:cross-view-matching}
\end{equation}

A lower cost indicates that the two meshes are more
likely to represent the same person.
A pair $(m_i,m_j)$  is accepted if the meshes mutually minimize the cost below a tunable threshold $\delta_{\mathrm{p2p}}$, \ie
    \begin{equation}
    \label{eq:mutual-argmax}
    \begin{cases}
    m_j = \operatorname*{arg\,min}_{m \in M_t^b} \mathcal{C}_{\mathrm{p2p}}(m_i,m), \\
    m_i = \operatorname*{arg\,min}_{m \in M_t^a} \mathcal{C}_{\mathrm{p2p}}(m,m_j), \\
    \mathcal{C}_{\mathrm{p2p}}(m_i,m_j) < \delta_{\mathrm{p2p}}.
    \end{cases}
    \end{equation}

A mesh remains unmatched if no candidate satisfies all three conditions in \cref{eq:mutual-argmax}. This allows observations that are visible in only one view to remain unpaired.

We compute $\mathcal{C}_{\mathrm{p2p}}$ as a weighted sum of three geometric errors:
\begin{equation}
\mathcal{C}_{\mathrm{p2p}}(m_i,m_j)
=
\lambda_J E_J
+\lambda_H E_H
+\lambda_O E_O,
\label{eq:p2p-cost}
\end{equation}
where $\lambda_J$, $\lambda_H$ and $\lambda_O$ are non-negative weights.
The joint error $E_J$ is the mean distance between the 17 COCO keypoints of $m_i$ and $m_j$
after optimal $\mathrm{Sim}(3)$ alignment. The head-position error $E_H$ is the world-frame distance between the two heads.
The head-orientation error $E_O$ is the angle between the two head-facing directions,
computed from the face and ear keypoints in the world frame too.

\noindent\textbf{Matching across time.}
We subsequently link meshes across time.
Specifically, meshes in $M_t^v$ are matched with those in
$M_{t+1}^v$ using the same person-to-person matching cost
$\mathcal{C}_{\mathrm{p2p}}$, with separate weights for temporal matching,
and the same rule as \cref{eq:mutual-argmax}.
Meshes without a valid match remain unmatched. 

\noindent\textbf{Grouping meshes.}
The accepted pairs from both matching steps are pooled
and grouped greedily.
Starting with each mesh as a singleton group, 
we process pairs in ascending order of $\mathcal{C}_{\mathrm{p2p}}$
and merge their groups, 
and skip any merge that would place two meshes from the same view and timestep in the same group.
Each resulting group collects meshes of the same person across views and time, while unmatched meshes remain singletons.

Together, these steps associate meshes of the same physical
person across views and time.
It remains to determine which egocentric camera belongs
to each group.

\noindent\textbf{Associating cameras to meshes.}
In the EgoMocap setting, each egocentric camera is mounted
on its wearer's forehead.
Therefore, if a mesh $m_i$ depicts the wearer of camera $v$,
its head position and orientation should agree with those
of camera $v$ at the same timestep.

Analogous to \cref{eq:cross-view-matching},
we define a person-to-camera matching cost
\begin{equation}
\mathcal{C}_{\mathrm{p2c}}:
M_t \times \{1,\ldots,V\}
\rightarrow \mathbb{R},\;
(m_i,v)
\mapsto
\mathcal{C}_{\mathrm{p2c}}(m_i,v),
\label{eq:anchor-likelihood}
\end{equation}
computed only for $v\neq\mathcal{V}(m_i)$, as a person is assumed never visible in their own camera.

We compute $\mathcal{C}_{\mathrm{p2c}}$ as a weighted sum
of three geometric errors:
\begin{equation}
\mathcal{C}_{\mathrm{p2c}}(m_i,v)
=
\lambda_P E_P
+\lambda_D E_D
+\lambda_B E_B,
\label{eq:p2c-cost}
\end{equation}
where $\lambda_P$, $\lambda_D$ and $\lambda_B$ are non-negative weights.
The position error $E_P$ is the world-frame distance
between the estimated head and camera $v$.
The orientation error $E_D$ is the angle between
the head-facing direction and the optical axis of camera $v$.
The bearing error $E_B$ is the angle between the directions
from the observing camera to the estimated head
and to camera $v$---independent of the estimated
head depth.

Since all meshes in a group depict the same person,
they must be associated with a single camera.
We therefore determine the association at the group level
by selecting the camera with the lowest average cost:
\begin{equation}
v^*(G)
=
\operatorname*{arg\,min}_{v\notin\mathcal{V}(G)}
\frac{1}{|G|}
\sum_{m_i\in G}
\mathcal{C}_{\mathrm{p2c}}(m_i,v),
\label{eq:camera-assignment}
\end{equation}
where $G$ denotes the set of meshes in the group 
and $\mathcal{V}(G)$ denotes the set of views that observed at least one mesh in $G$.
The constraint $v \notin \mathcal{V}(G)$ excludes the observing views as candidate cameras.

Finally, a group may correspond to a non-participant,
\ie a passerby who does not wear an egocentric camera.
We therefore accept $v^*(G)$ only when its average cost
is below a tunable threshold $\delta_{\mathrm{p2c}}$.
Otherwise, the group is classified as a non-participant
and excluded from the output of \method.
If multiple groups select the same camera,
only the group with the lowest average cost
keeps the association.
As a result, each camera is associated with at most
one group.
We denote the group associated with camera $v$ by $G^v$,
which collects the meshes of participant $v$.

\subsection{DIFFUSION-GUIDED MOTION COMPLETION}
\label{sec:method:diffusion}

\begin{figure}[t]
  \centering
  \includegraphics[width=\linewidth]{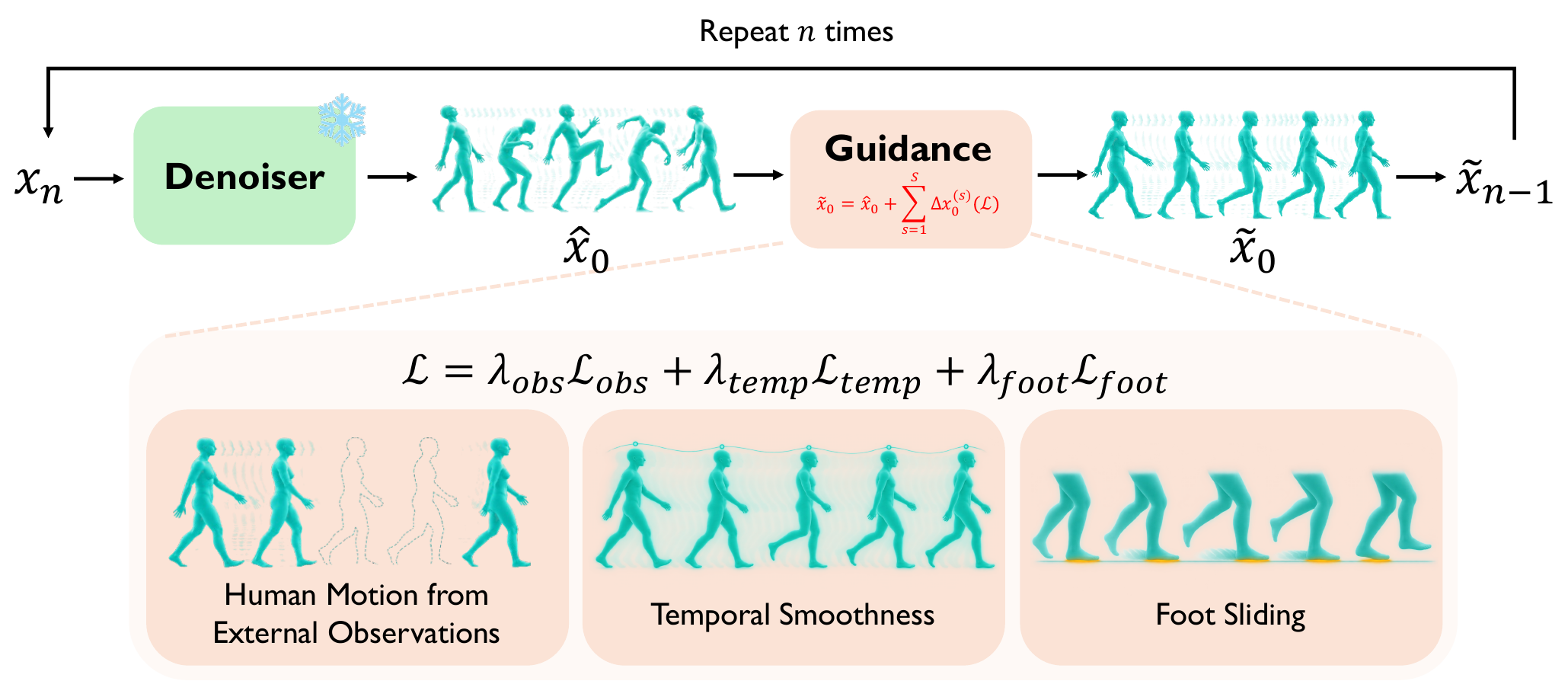}
  \caption{Diffusion-guided motion completion.
    At each denoising step $n$, the current noisy motion $\boldsymbol{x}_n$ is passed to
    the frozen denoiser (EgoAllo), which attempts to predict the clean motion
    $\hat{\boldsymbol{x}}_0$.
    The external-view observations guide the correction of $\hat{\boldsymbol{x}}_0$
    into $\tilde{\boldsymbol{x}}_0$, from which the next noisy state
    $\tilde{\boldsymbol{x}}_{n-1}$ is computed and fed back to the denoiser.}
  \label{fig:diffusion}
\end{figure}

The associated meshes from external views provide body pose
observations only at timesteps where a participant is visible.
To reconstruct complete motion,
we use EgoAllo~\cite{yi2025egoallo}, a pretrained motion diffusion
model conditioned on the head trajectory, and guide its
generation using the obtained pose estimations.
For participant $v$, the head trajectory is obtained from the camera poses 
estimated in \cref{sec:method:backbone},
chained over all chunks into $\{\boldsymbol{\pi}_t^v\}_{t=1\ldots T}$.
Since EgoAllo is based on SMPL-H~\cite{romero2017mano},
we convert its output to SMPL-X.
\cref{fig:diffusion} illustrates this guided denoising process.

\noindent\textbf{Guided DDIM sampling.}
At each denoising step $n$, EgoAllo attempts to predict the clean motion
$\hat{\boldsymbol{x}}_0$.
The deterministic DDIM~\cite{song2020denoising} update is given by
\begin{equation}
\boldsymbol{x}_{n-1}
=
\sqrt{\bar{\alpha}_{n-1}}\,
\hat{\boldsymbol{x}}_0
+
\sqrt{1-\bar{\alpha}_{n-1}}\,
\hat{\boldsymbol{\epsilon}}_n,
\label{eq:ddim}
\end{equation}
where $\bar{\alpha}_n$ denotes the cumulative noise schedule
and $\hat{\boldsymbol{\epsilon}}_n$ is the estimated noise.
To incorporate external observations during sampling,
we iteratively correct the predicted clean motion
by minimizing a guidance loss $\mathcal{L}$:
\begin{equation}
\tilde{\boldsymbol{x}}_0
=
\hat{\boldsymbol{x}}_0
+
\sum_{s=1}^{S}
\Delta\boldsymbol{x}_0^{(s)}(\mathcal{L}),
\label{eq:guidance-update}
\end{equation}
where $\Delta\boldsymbol{x}_0^{(s)}(\mathcal{L})$
denotes the update at the $s$-th of $S$
gradient-based optimization steps, which start
from $\hat{\boldsymbol{x}}_0$.
Replacing $\hat{\boldsymbol{x}}_0$ with
$\tilde{\boldsymbol{x}}_0$ in \cref{eq:ddim}, while keeping
the noise estimate fixed, yields
\begin{equation}
\begin{aligned}
\tilde{\boldsymbol{x}}_{n-1}
&=
\sqrt{\bar{\alpha}_{n-1}}\,
\tilde{\boldsymbol{x}}_0
+
\sqrt{1-\bar{\alpha}_{n-1}}\,
\hat{\boldsymbol{\epsilon}}_n\\
&=
\boldsymbol{x}_{n-1}
+
\sqrt{\bar{\alpha}_{n-1}}\,
\sum_{s=1}^{S}
\Delta\boldsymbol{x}_0^{(s)}(\mathcal{L}).
\end{aligned}
\label{eq:guidance-state}
\end{equation}
Unlike standard DDIM,
the guided update incorporates
observation-driven corrections into the next noisy state.
These corrections therefore influence subsequent denoising
steps, rather than modifying only the final motion.
The pretrained diffusion model remains frozen.

\noindent\textbf{Guidance loss.}
We define the guidance loss as
\begin{equation}
\mathcal{L}
=
\lambda_{\mathrm{obs}}\mathcal{L}_{\mathrm{obs}}
+
\lambda_{\mathrm{temp}}\mathcal{L}_{\mathrm{temp}}
+
\lambda_{\mathrm{foot}}\mathcal{L}_{\mathrm{foot}}.
\label{eq:guidance-loss}
\end{equation}
Here, $\mathcal{L}_{\mathrm{obs}}$ aligns the generated motion
with external-view observations,
$\mathcal{L}_{\mathrm{temp}}$ penalizes joint acceleration
to encourage temporal smoothness, and
$\mathcal{L}_{\mathrm{foot}}$ penalizes foot sliding
during contact.
The loss $\mathcal{L}$ is computed independently
for each participant $v$.

Let $\mathcal{J}_t^v \in \mathbb{R}^{K\times3}$
denote the world-frame joint positions of participant $v$
at timestep $t$ in the predicted clean motion, where
$K$ is the number of joints.
We denote the corresponding head-relative joint positions
by $\bar{\mathcal{J}}_t^v$,
and use a subscript $k$ for the $k$-th joint, \ie $\bar{\mathcal{J}}_{t,k}^v \in \mathbb{R}^{3}$.
Similarly, $\mathcal{J}(m_i)$ and
$\bar{\mathcal{J}}(m_i)$ denote the world-frame and
head-relative joint positions of an observed mesh $m_i$,
respectively.
We use $G_t^v$ to denote the set of observed meshes
associated with participant $v$ at timestep $t$.

The observation loss measures the difference between
generated and observed head-relative joint positions:
\begin{equation}
\mathcal{L}_{\mathrm{obs}}
=
\frac{1}{Z}
\sum_{t=1}^{T}
\sum_{m_i\in G_t^v}
\sum_{k=1}^{K}
\rho\left(
\left\|
\bar{\mathcal{J}}_{t,k}^v
-
\bar{\mathcal{J}}_k(m_i)
\right\|_2
\right),
\label{eq:guidance-obs}
\end{equation}
where $\rho$ is the Huber penalty and
$Z$ counts the total number of joints over all observed meshes, $Z = K \sum_{t = 1}^{T} |G_t^v|$.
This loss is evaluated only where observations exist.

To encourage temporal consistency, we penalize
head-relative joint acceleration:
\begin{equation}
\mathcal{L}_{\mathrm{temp}}
=
\frac{1}{(T-2)K}
\sum_{t=2}^{T-1}
\sum_{k=1}^{K}
\left\|
\frac{
\bar{\mathcal{J}}_{t+1,k}^{v}
-2\bar{\mathcal{J}}_{t,k}^{v}
+\bar{\mathcal{J}}_{t-1,k}^{v}
}{\Delta t^2}
\right\|_2^2.
\label{eq:guidance-temp}
\end{equation}

We further penalize foot sliding during predicted contact
using world-frame foot velocities:
\begin{equation}
\mathcal{L}_{\mathrm{foot}}
=
\frac{1}{(T-1)F}
\sum_{t=1}^{T-1}
\sum_{f=1}^{F}
c_{t,f}^v
\left\|
\frac{
\mathcal{J}_{t+1,f}^v
-\mathcal{J}_{t,f}^v
}{\Delta t}
\right\|_2^2,
\label{eq:guidance-foot}
\end{equation}
where $f$ indexes the $F$ foot joints,
$\mathcal{J}_{t,f}^v$ denotes the world-frame position of foot joint $f$,
$c_{t,f}^v$ denotes its contact
probability predicted by EgoAllo and treated as a constant,
and $\Delta t$ is the frame interval.

Together, these losses guide the motion toward external observations while preserving temporal consistency at unobserved timesteps and enhancing physical plausibility.

\section{EXPERIMENTS}
\label{sec:experiments}

\subsection{EXPERIMENTAL SETUP}
\label{sec:exp:setup}

\begin{table*}[t]
  \centering
  \small
  \setlength{\tabcolsep}{5pt}
  \caption{
    Pose accuracy and motion quality on 34 EgoHumans sequences.
    Pose accuracy 
    evaluates 3D joint errors,
    while motion quality 
    evaluates temporal smoothness and physical plausibility.
    ReID + Interpolation yields the lowest \pampjpe for per-frame local poses,
    but degrades on sequence-level metrics and produces jittery motion. 
    In contrast, \method achieves the best sequence-level accuracy and motion quality.
    Best values are in bold.
  }
  \begin{tabular}{@{}l ccc ccc@{}}
    \toprule
    & \multicolumn{3}{c}{\textbf{Pose Accuracy}}
    & \multicolumn{3}{c}{\textbf{Motion Quality}} \\
    \cmidrule(lr){2-4} \cmidrule(l){5-7}
    Method
    & \pampjpe$\downarrow$
    & \wampjpe$\downarrow$
    & \rampjpe$\downarrow$
    & Jitter$\downarrow$
    & Accel$\downarrow$
    & Foot Sliding$\downarrow$ \\
    \midrule
    Pi3X~\cite{pi3_2026} + T-Pose
    & 193.0 & 281.7 & 402.4
    & 0.290 & 29.8 & 24.0 \\
    Pi3X~\cite{pi3_2026} + EgoAllo~\cite{yi2025egoallo}
    & 150.9 & 220.7 & 258.3
    & 0.278 & 31.2 & 24.7 \\
    \midrule
    ReID + Interpolation
    & \textbf{64.8} & 203.4 & 298.9
    & 1.208 & 88.4 & 63.2 \\
    ReID + Unguided EgoAllo
    & 76.3 & 213.0 & 307.4
    & 1.328 & 96.8 & 71.5 \\
    \method (Ours)
    & 77.6 & \textbf{145.5} & \textbf{198.1}
    & \textbf{0.241} & \textbf{26.8} & \textbf{23.9} \\
    \bottomrule
  \end{tabular}
  \label{tab:pose_comparison}
\end{table*}

\begin{figure*}[t]
  \centering
  \includegraphics[width=\linewidth]{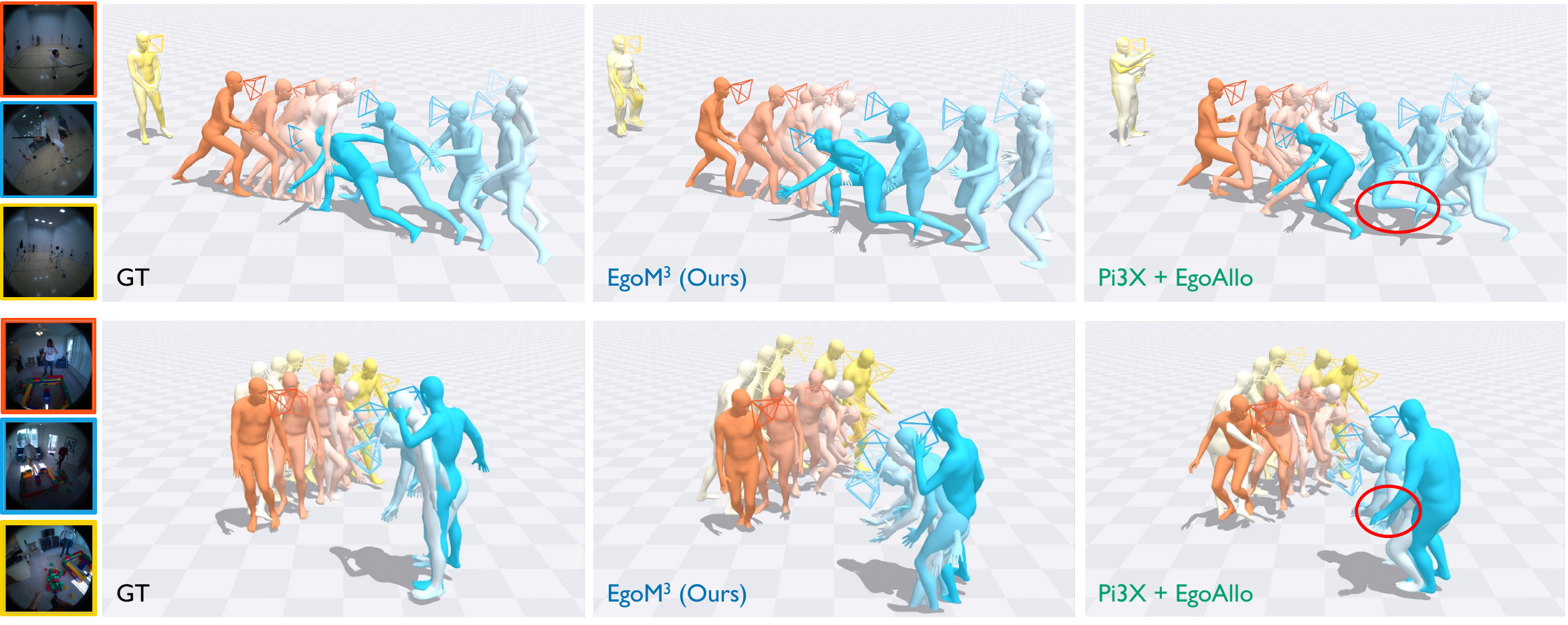}
  \caption{Qualitative results of global multi-person motion reconstruction. We compare \method with Pi3X + EgoAllo and show ground truth. Human meshes at multiple timesteps are overlaid in a shared world coordinate system, with different colors representing different individuals. \method reconstructs coherent multi-person motion and preserves the relative spatial arrangement of individuals.
  Top: fencing; Bottom: Lego assembly. Left: the egocentric views, framed in the color of their wearer.
  Red circles mark noticeable reconstruction errors of the Pi3X + EgoAllo baseline.}
  \label{fig:qualitative}
\end{figure*}

\begin{figure*}[t]
  \centering
  \includegraphics[width=\linewidth]{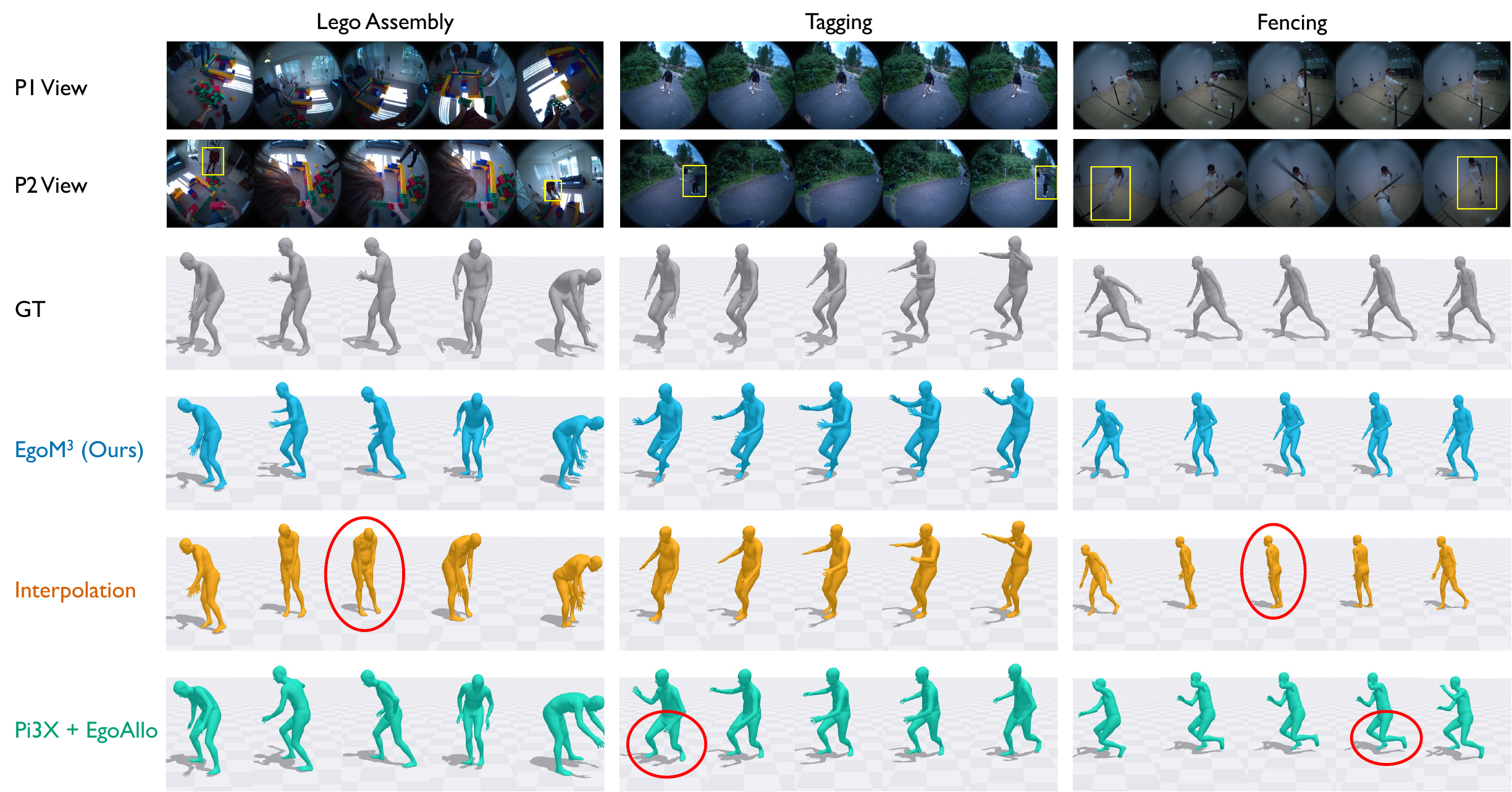}
    \vspace{-8pt}
  \caption{Motion completion when the human is out of view (Lego Assembly and Tagging) or human stream detection fails (Fencing).
    All the human meshes in this figure correspond to P1.
    In P2 View, P1 is only available at the first and fifth frames.
    Yellow boxes indicate frames in which P1 is visible and detected.
    With diffusion guidance, \method produces coherent human meshes that closely follow the ground truth,
    while the other two methods exhibit noticeable reconstruction errors.
    Red circles mark these errors.
    Note that reconstruction is performed on the full sequences, and only short excerpts are presented here.
  }
  \label{fig:qualitative_undetected}
\end{figure*}

\noindent\textbf{Dataset.}
We evaluate \method on EgoHumans~\cite{khirodkar2023egohumans}, a multi-person dataset comprising 133 sequences across seven activity scenarios. Each sequence features 2--4 participants equipped with egocentric cameras. Ground-truth annotations were obtained using an additional 8--15 exocentric cameras. Following the standard EgoHumans training and evaluation split, we train on four scenarios (basketball, volleyball, badminton, and tennis) and evaluate on the remaining three scenarios (tagging, Lego assembly, and fencing), totaling 34 held-out sequences.

\noindent\textbf{Implementation details.}
We train the head detector, the fusion MLP and the output heads jointly for 20 epochs on $T_c{=}8$ timesteps, while keeping Pi3X and Multi-HMR frozen.
We use the AdamW optimizer~\cite{loshchilov2019adamw} with a learning rate of $10^{-4}$, a weight decay of $10^{-4}$, and a batch size of four chunks; the learning rate follows a linear warm-up followed by cosine decay.
The ReID cost weights and thresholds are determined through hyperparameter search.
For~\cref{eq:guidance-loss}, we set $\lambda_{\mathrm{obs}}{=}1$ and $\lambda_{\mathrm{temp}}{=}\lambda_{\mathrm{foot}}{=}0.1$, and run $S{=}25$ Adam~\cite{kingma2015adam} iterations with a step size of $0.03$ at each denoising step.
We also use $T_c{=}8$ during inference.

\noindent\textbf{Evaluation metrics.}
We evaluate pose accuracy and motion quality.
For pose accuracy, we use three metrics to measure the mean position error of the 3D joints after performing 
a $\mathrm{Sim}(3)$ alignment to the ground truth:
i) \pampjpe~\cite{kim2026harmoni} aligns each person in each frame separately, removing the global position, orientation, and scale, and thus evaluates the local body pose per frame.
ii) \wampjpe~\cite{shin2024wham} aligns each person once per 100-frame segment, using all joints of all frames in the segment, and thus evaluates the motion of each individual, including the global trajectory.
iii) \rampjpe is a new group-level metric that aligns all people jointly with a single transformation per 100-frame segment. 
The alignment is performed jointly over all joints, all frames and all people. 
Since the alignment transformation is shared by the whole participant group and across time, \rampjpe measures the spatiotemporal pose accuracy at the group level.
All three pose metrics are measured in millimeters.

For motion quality, Jitter measures the mean third finite difference of the predicted joint positions in $\mathrm{\text{km}/\text{s}^3}$, Accel measures the error of the joint accelerations against the ground truth in mm/frame$^2$, and Foot Sliding measures the horizontal displacement of the ankles while the ground-truth foot is in contact with the floor, in mm/frame.

\subsection{RESULTS}
\label{sec:exp:comparison}

We compare to two baselines that use the camera trajectories estimated by Pi3X~\cite{pi3_2026}.
First the baseline \emph{Pi3X + T-Pose} places an upright T-pose at each wearer's camera, facing its heading.
This naive baseline demonstrates the diverse poses of the evaluated sequences.
The second baseline \emph{Pi3X + EgoAllo} conditions EgoAllo~\cite{yi2025egoallo} on the camera trajectories,
\ie estimating each wearer from their own camera trajectory.

Apart from the above two baselines, we compare against two variants of \method that replace diffusion guidance with alternative motion completion strategies after the ReID stage (\cref{sec:method:assoc}).
Both variants keep the view-averaged mesh estimates on observed frames and differ in how unobserved frames are filled:
i) \emph{ReID + Interpolation} linearly interpolates between neighboring observed poses, and
ii) \emph{ReID + Unguided EgoAllo} fills unobserved frames using the output of EgoAllo without guidance.

\cref{tab:pose_comparison} shows the quantitative comparison results.
ReID + Interpolation achieves the best per-frame local pose (64.8\,mm \pampjpe), but has lower sequence-level accuracy---higher \wampjpe and \rampjpe---and much higher jitter than \method.
In contrast, \method achieves the best sequential accuracy at both the individual level (145.5\,mm \wampjpe) and the group level (198.1\,mm \rampjpe), as well as the best motion quality across all three metrics.
ReID + Unguided EgoAllo performs worse than ReID + Interpolation on all metrics, and worse than \method on all metrics except \pampjpe.
The two Pi3X-only baselines fail to produce accurate local poses 
since they rely only on camera trajectories without observing the wearers' bodies.

\cref{fig:qualitative} shows qualitative results of global multi-person motion reconstruction.
\method reconstructs coherent motion of all participants in a shared world coordinate system and preserves their relative arrangement, whereas Pi3X + EgoAllo shows noticeable errors.
\cref{fig:qualitative_undetected} further shows motion completion for a person who is observed from the external views in the first and last frames but unavailable in all frames in between.
In these unavailable frames, \method produces more coherent motion than ReID + Interpolation and Pi3X + EgoAllo.

\begin{table}[t]
  \centering
  \small
  \setlength{\tabcolsep}{3pt}
  \caption{Ablation study of the guidance loss terms in
  \cref{eq:guidance-loss}.
  Each variant removes one loss term while keeping the others
  unchanged.
  PA, WA, and RA denote \pampjpe, \wampjpe,
  and \rampjpe, respectively.
  The best values are highlighted in bold.}
  \begin{tabular}{@{}lccccc@{}}
      \toprule
      Variant
      & PA$\downarrow$
      & WA$\downarrow$
      & RA$\downarrow$
      & Jitter$\downarrow$
      & Foot Sliding$\downarrow$ \\
      \midrule
      \method
      & 77.6
      & \textbf{145.5}
      & \textbf{198.1}
      & 0.241
      & \textbf{23.9} \\
      w/o observation term $\mathcal{L}_{\mathrm{obs}}$
      & 142.6 & 212.5 & 269.1 & \textbf{0.234} & 24.7 \\
      w/o temporal term $\mathcal{L}_{\mathrm{temp}}$
      & \textbf{77.0} & 146.3 & 198.7 & 0.292 & 27.5 \\
      w/o foot contact term $\mathcal{L}_{\mathrm{foot}}$
      & 77.5 & 148.3 & 201.7 & 0.240 & 30.6 \\
      \bottomrule
    \end{tabular}
  \label{tab:ablation-guidance}
\end{table}

\noindent\textbf{Ablation Study on Diffusion Guidance.}
To analyze the diffusion guidance, we compare the full model with variants that each remove one term of the guidance loss, as shown in \cref{tab:ablation-guidance}.
Removing the observation term slightly reduces the jitter, but worsens \pampjpe by 84\% and \wampjpe by 46\%, since the motion is then conditioned only on the wearer's head trajectory.
Removing the temporal term barely changes the overall accuracy but increases the jitter by 21\% and the foot sliding by 15\%.
Removing the foot contact term increases the foot sliding by 28\%.
These results show that the three terms complement each other.
The observation term makes the motion follow the per-view observations, while the temporal and foot contact terms keep it smooth and physically plausible.

\section{CONCLUSION}
\label{sec:conclusion}

We study EgoMocap, the problem of reconstructing a group of people's
3D body meshes from their egocentric RGB videos.
To tackle EgoMocap,
we propose \method, a three-stage method combining
a dual-encoder backbone for per-view mesh and camera poses,
a geometry-based ReID process across views and time, and
diffusion-guided motion completion for missing observations.
Experiments on EgoHumans demonstrate that \method
improves global pose accuracy and motion quality
over existing baselines.

\noindent\textbf{Limitations and future work.}
In \method, motion completion may fail during
prolonged periods of missing observations, as the head
trajectory and motion prior alone are insufficient for detailed limb motions.
\method further relies on existing modules for camera
pose estimation, whose errors can propagate
to geometry-based association.
Regarding the evaluation,
we used the \rampjpe metric to measure participants' relative spatial arrangement, but it does not explicitly
assess human--human interactions, such as one participant
tapping another's shoulder. 
In addition, our evaluation is limited to a single dataset
with up to four participants.
Expanding the EgoMocap setting to sequences with many more participants,
and evaluating fine-grained interactions, are left for future work.

\section*{ACKNOWLEDGMENTS}
This work was partly supported by JST BOOST, Japan Grant Number JPMJBS2409. The authors would like to thank the members of the Saito Laboratory for their insightful discussions and valuable feedback on the research direction and methodology.
Z. Zhu also thanks Prof. Yoichi Sato for the invitation to The University of Tokyo that initiated this collaboration.

\bibliographystyle{IEEEtran}
\bibliography{main}

\end{document}